**Title**

A deep neural network to enhance prediction of 1-year mortality using echocardiographic videos of the heart

**Authors**

Alvaro Ulloa, MS[abf]; Linyuan Jing, PhD[ab]; Christopher W Good, DO[d];

David P vanMaanen, MS[ab]; Sushravya Raghunath, PhD[ab]; Jonathan D Suever, PhD[ab];

Christopher D Nevius, BT[ab]; Gregory J Wehner, PhD[c]; Dustin Hartzel, BS[b]; Joseph B Leader, BA[b]; Amro Alsaid, MBBCh[d]; Aalpen A Patel, MD[e]; H Lester Kirchner, PhD[b]; Marios S Pattichis[f], PhD; Christopher M Haggerty, PhD[ab]; Brandon K Fornwalt, MD PhD[abde]

[a]Dept of Imaging Science and Innovation, Geisinger, Danville, PA, United States

[b]Biomedical and Translational Informatics Institute, Geisinger, Danville, PA, United States

[c]Dept of Biomedical Engineering, University of Kentucky, Lexington, KY, United States

[d]Heart Institute, Geisinger, Danville, PA, United States

[e]Dept of Radiology, Geisinger, Danville, PA, United States

[f]Electrical and Engineering Dept, University of New Mexico, Albuquerque, NM, United States

**Authors Contributions.** A.U., C.H., and B.F conceived the study and designed the experiments. A.U. conducted all experiments. A.U. and S.R. wrote the software for applying deep learning to Echocardiography videos. A.U., L.J., D.V., D.H., J.S., and J.L. assembled the input data. L.K., G.W., M.P., and A.P. gave advice on experiment design. L.J., C.N., C.H., and B.F. manually audited the data for the cardiologist survey. C.G. and A.A. completed the survey. A.U., C.H., M.P., and B.F wrote the manuscript. All authors critically revised the manuscript.

**Author Information.** No conflicts of interest were disclosed. Correspondence and requests for materials should be addressed to bkf@gatech.edu

**Contact Information:**

AU = alvarouc@gmail.com

LJ = ljing@geisinger.edu

CG = cwgood1@geisinger.edu

DV = dpvanmaanen@geisinger.edu

SR = sraghunath@geisinger.edu

JS = jdsuever@geisinger.edu

CN = cdnevius@geisinger.edu

GW = wehnergj@gmail.com

DH = dnhartzel@geisinger.edu

JL = jbleader@geisinger.edu

AA = aalsaid@geisinger.edu

AP = aapatel@geisinger.edu

HK = hlkirchner@geisinger.edu

MP = pattichi@unm.edu

CH = cmhaggerty@geisinger.edu

BF = bkf@gatech.edu

**Abstract**

Predicting future clinical events helps physicians guide appropriate intervention. Machine learning has tremendous promise to assist physicians with predictions based on the discovery of complex patterns from historical data, such as large, longitudinal electronic health records (EHR). This study is a first attempt to demonstrate such capabilities using raw echocardiographic videos of the heart. We show that a large dataset of 723,754 clinically-acquired echocardiographic videos (~45 million images) linked to longitudinal follow-up data in 27,028 patients can be used to train a deep neural network to predict 1-year mortality with good accuracy (area under the curve (AUC) in an independent test set = 0.839). Prediction accuracy was further improved by adding EHR data (AUC = 0.858). Finally, we demonstrate that the trained neural network was more accurate in mortality prediction than two expert cardiologists. These results highlight the potential of neural networks to add new power to clinical predictions.

**Introduction**

Imaging is critical to treatment decisions in most modern medical specialties and has also become one of the most data rich components of electronic health records (EHRs). For example, during a single routine ultrasound of the heart (an echocardiogram), approximately 10-50 videos (~3,000 images) are acquired to assess heart anatomy and function. In clinical practice, a cardiologist realistically has 10-20 minutes to interpret these 3,000 images within the context of numerous other data streams such as laboratory values, vital signs, additional imaging studies (radiography, magnetic resonance imaging, nuclear imaging, computed tomography) and other diagnostics (e.g. electrocardiogram). While these numerous sources of data offer the potential for more precise and accurate clinical predictions, humans have limited capacity for data integration in decision making.[1] Hence, there is both a need and a substantial opportunity to leverage technology, such as artificial intelligence and machine learning, to manage this abundance of data and ultimately provide intelligent computer assistance to physicians.[2,3]

Recent advances in "deep" learning (deep neural network; DNN) technologies; such as Convolutional Neural Networks (CNNs), Recurrent Neural Networks (RNN), Dropout Regularization, and adaptive gradient descent algorithms[4]; in conjunction with massively parallel computational hardware (graphic processing units), have enabled state-of-the-art predictive models for image, time-series, and video-based data[5,6]. For example, DNNs have shown promise in diagnostic applications, such as diabetic retinopathy[7], skin cancer[8], pulmonary nodules[9], cerebral microhemorrhage[10,11], and etiologies of cardiac hypertrophy[12]. Yet, the opportunities with machine learning are not limited to such diagnostic tasks.[2]

Prediction of future clinical events, for example, is a natural but relatively unexplored extension of machine learning in medicine. Nearly all medical decisions rely on accurate prediction. A diagnosis is provided to patients since it helps to establish the typical future clinical course of patients with similar symptoms, and a treatment is provided as a prediction of how to positively impact that predicted future clinical course. Thus, using computer-based methods to directly predict future clinical events is an important task where computers can likely assist human interpretation due to the inherent complexity of this problem. For example, a recent article in 216,221 patients demonstrated how a Random Forest model can predict in-hospital mortality with high accuracy[13]. Deep learning models have also recently been used to predict mortality risk among hospitalized patients to assist with palliative care referrals.[14] In cardiology, variables derived from electronic health records have been used to predict two-to-five year all-cause mortality in patients undergoing coronary computed tomography[15,16], five-year cardiovascular mortality in a general clinical population[17], and up to five-year all-cause mortality in patients undergoing echocardiography[18].

Notably, these initial outcome prediction studies in cardiology exclusively used human-derived, i.e. "hand-crafted" features from imaging, as opposed to automatically analyzing the raw image data. While this use of hand-crafted features is important, an approach that is unbiased by human opinions and not limited by human perception, human ability in pattern recognition, and effort may be more robust. That is, there is strong potential in an automated analysis that would leverage all available data in the images rather than a few selected clinical or clinically inspired measurements. Furthermore, the potential benefit of this approach for echocardiography may be enhanced by the added availability of rich temporal (video) data. DNNs make this unique approach possible. However, using video data also increases technical complexity and thus initial efforts to apply deep learning to echocardiography have focused on ingesting individual images rather than full videos.[19]

In this paper, we show that a DNN can predict 1-year mortality directly from echocardiographic videos with good accuracy and that this accuracy can be improved by incorporating additional clinical variables from the electronic health record. We do this through a technical advance that leverages the full echocardiographic videos to make predictions using a three-dimensional DNN. In addition to this technical advance, we demonstrate direct clinical relevance by showing that the DNN is more accurate in predicting 1-year mortality compared to two expert physician cardiologists.

**Results**

We utilize a fully 3D Convolutional Neural Network (CNN) design in this study (Figure 1). CNNs are neural networks that exploit spatial coherence in an image to significantly reduce the number of parameters that a fully connected network would need to learn. CNNs have shown promise in image classification tasks[4], even surpassing human abilities[20]. Details of additional model architectures attempted (including a time-distributed 2D CNN + long short term memory network [LSTM][21–24]) are described in the methods.

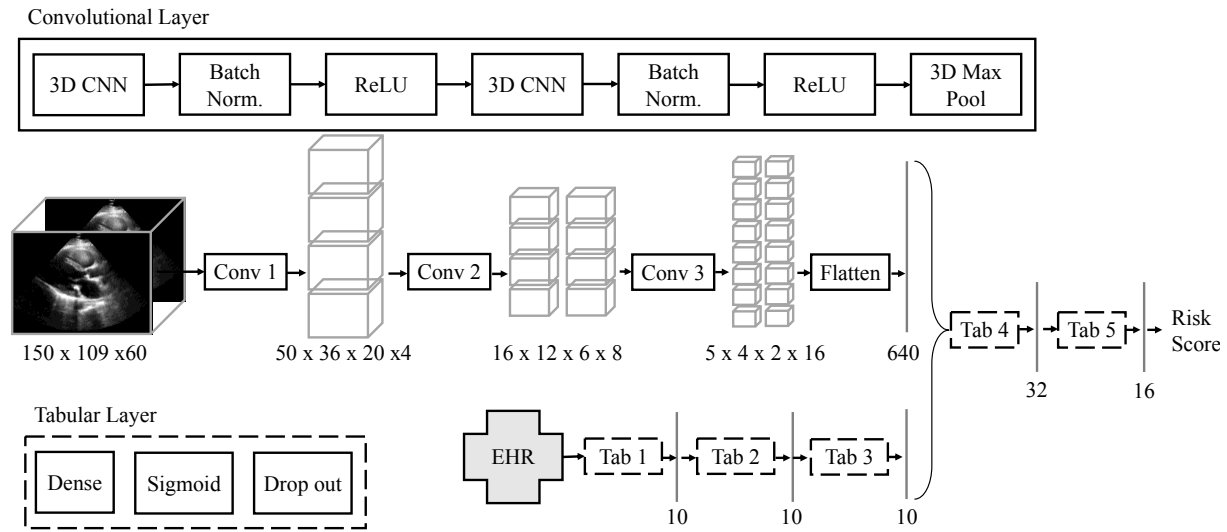

*Figure 1: Neural network architecture for mortality prediction from echocardiography videos and electronic health record (EHR) data. The convolutional layer (Conv) is shown on the top box with a solid outline and the tabular layer (Tab) is shown in the bottom box with a dashed outline. The convolutional layer consists of Convolutional Neural Networks (CNN), Batch Normalizations (Batch Norm.), rectified linear units (ReLU), and a three-dimensional Maximum*

*Pooling layer (3D Max Pool). The tabular layer consists of a fully connected layer (Dense) with sigmoid activations and a Drop Out layer. The input video dimensions were 150 x 109 x 60 pixels, and the output dimension of every layer are shown.*

We first collected 723,754 clinically acquired echocardiographic videos (approximately 45 million images) from 27,028 patients that were linked to at least 1 year of longitudinal follow-up data to know whether the patient was alive or dead within that time frame. Overall, 16% of patients in this cohort were deceased within a year after the echocardiogram was acquired. Based on a power calculation detailed in the methods, we separated data from 600 patients for validation and comparison against two independent cardiologists and used the remaining data for 5-fold cross-validation schemes.

During the acquisition of an echocardiogram, images of the heart and large blood vessels are acquired in different two-dimensional planes, or "views", that are standardized according to clinical guidelines[25]. We generated separate models for each of the 21 standard echocardiographic views and showed that the proposed models were able to accurately predict 1-year survival using only the raw video data as inputs (Figure 2). The chosen 3D CNN architecture (AUC range: 0.695–0.784) outperformed the 2D CNN + LSTM architecture (AUC range: 0.703–0.752) for most views. In both cases, the parasternal long-axis ("PL DEEP") view had the best performance. This result was in line with clinical intuition, since the PL DEEP view is typically reported by cardiologists as the most informative "summary" view of overall cardiac health. This is because the PL DEEP view contains elements of the left ventricle, left atrium, right ventricle, aortic and mitral valves, and whether or not there is a pericardial or left pleural effusion all within a single view.

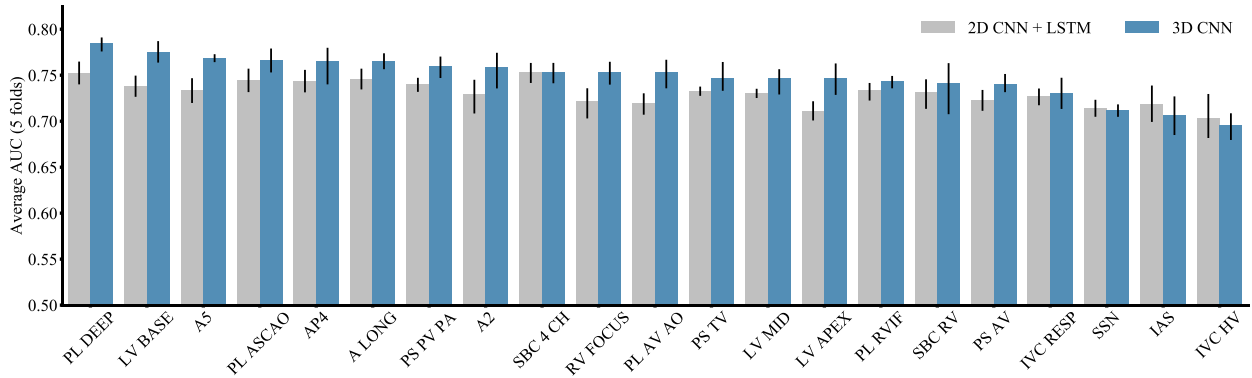

*Figure 2: One-year mortality prediction performance ranking for each echocardiography view alone (no EHR data) using the 2D CNN + LSTM architecture (gray) and 3D CNN (blue) models. The error bars denote one standard deviation above and below the average across 5 folds. See Extended Data Table 1 for all view label abbreviations.*

These results were relatively insensitive to image resolution (no significant difference was observed between models using full native resolution images (400 x 600 pixels) and reduced resolution images (100 x 150 pixels); Extended Data Figure 3). Similarly, adding derived optical flow velocity maps[26] to the models along with the pixel level data did not improve prediction accuracy (Extended Data Figure 4).

Next, we investigated the predictive accuracy of the models at additional survival intervals, including 3, 6, 9, and 12-month intervals after echocardiography. The models generally performed better at longer intervals, but AUCs for all cases were greater than 0.64 (Figure 3).

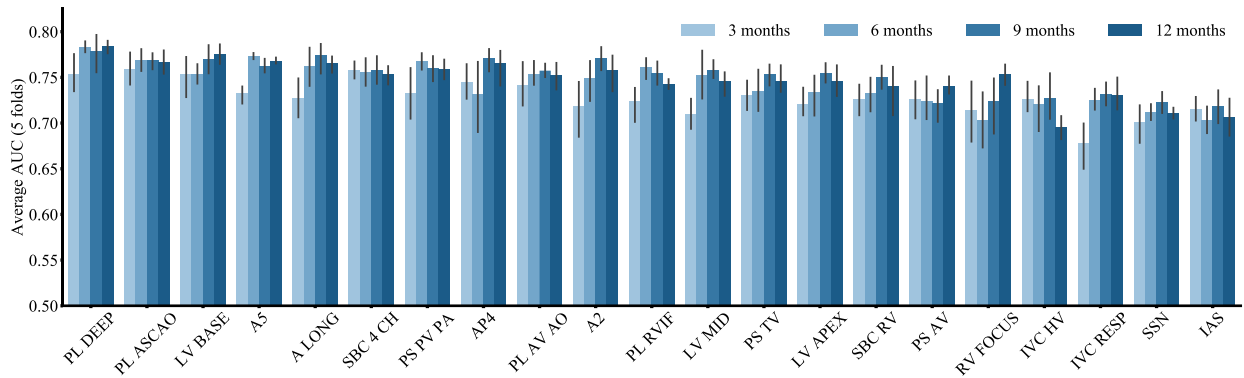

*Figure 3: Mortality prediction performance for echocardiographic videos alone at 3, 6, 9 and 12 months for all views. The error bars denote one standard deviation above and below the average across 5 folds.*

We then added select clinical ("EHR") variables from each patient including age, tricuspid regurgitation maximum velocity, heart rate, low density lipoprotein [LDL], left ventricular ejection fraction, diastolic pressure, pulmonary artery acceleration time, systolic pressure, pulmonary artery acceleration slope, and diastolic function. These 10 variables have previously been shown to contain >95% of the power for predicting 1-year survival in 171,510 patients[18] and their addition improved accuracy to predict 1-year survival for all echocardiographic views, with AUCs ranging from 0.79-0.82 (compared to 0.70-0.78 without these 10 EHR variables).

Next, we developed a software platform (see Methods) that we used to display an echocardiographic video of interest along with the 10 select EHR variables to two independent cardiologist echocardiographers who were blinded to the clinical outcomes. The cardiologists assessed whether each of 600 patients (independent test set extracted randomly from the original dataset of parasternal long axis views and not used for training of the machine) would be alive at one year based on the data presented. The final trained model (trained in all but these 600) was also applied to the same independent test set.

The overall accuracy of the model (75%) was significantly higher than that of the cardiologists (56% and 61%, p = 4.2 x $10^{-11}$ and 6.9 x $10^{-7}$ by Bonferroni-adjusted post-hoc analysis, Figure 4a. We found that the cardiologists tended to overestimate survival likelihood, yielding high specificities (97% and 91%, respectively) but poor sensitivities (16% and 31%, respectively) while the model, by design, balanced sensitivity and specificity (both 75%). Moreover, as demonstrated in Figure 4b, the operating points for the individual cardiologists fell within the envelope of the model's receiver operating characteristic curve (as opposed to falling at a different point on the same curve), suggesting inferior predictive performance in this task.

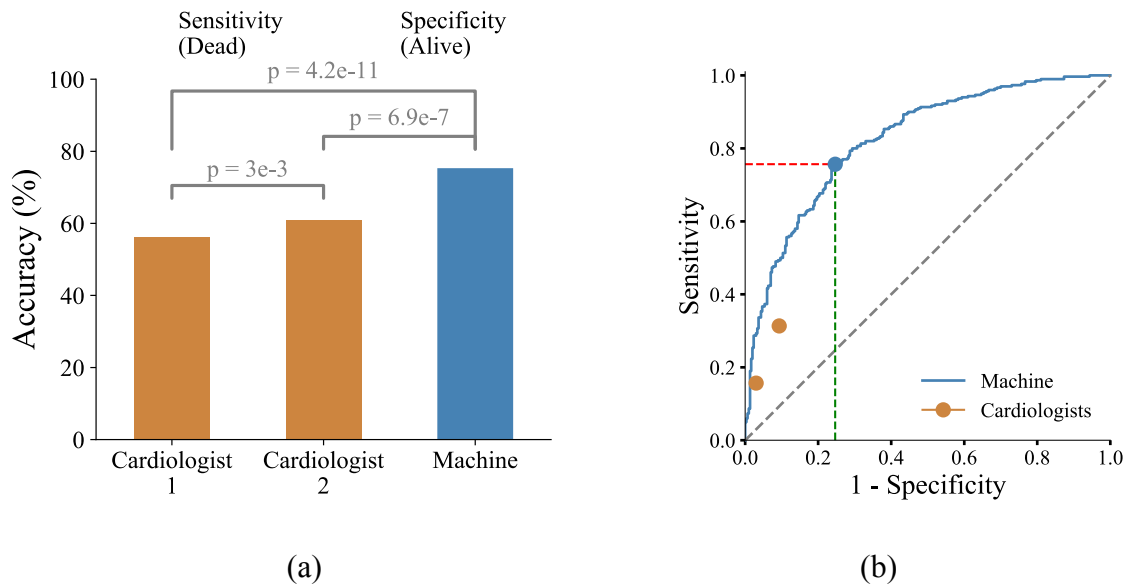

(a)                                           (b)

*Figure 4: Cardiologists vs Machine performance for 1-year mortality prediction from the survey dataset of 600 samples with balanced prevalence. The left plot (a) shows the accuracy in bars and sensitivity (red) and specificity (green) as triangles. The right plot (b) shows the operating*

*points of the cardiologists as orange dots, the Receiver Operating Characteristic curve for the machine performance in blue, and the machine operating point as a blue dot.*

Beyond the limited inputs selected for the clinical expert comparison, we sought to further characterize the model performance unconstrained by data input limitations. That is, we completed additional experiments permuting the input combinations of structured data (none, limited set [top 10 EHR variables], full set [158 EHR variables, as described in methods]) and echocardiography videos (none, single view, all 21 views). Models without videos were trained using all available data in our structured echocardiography measurement database (501,449 valid studies), while the models with videos were trained with all videos available for each view, ranging from 11,020 to 22,407 for single videos and 26,428 combined. In all cases, the test set was the 600 patients held out for the clinical expert comparison.

Table 1 shows that all videos combined with the full EHR variable set had the highest AUC in the held out test set of 600 studies, demonstrating the potential to further enhance the performance of the already clinically superior model. Several general trends were also noted. First, a single video view out-performed a model that included 10 EHR variables as input. Second, multiple videos had higher performance than single videos. Third, the learning curves (Figure 5) for multi-video predictions demonstrated that, despite having access to a massive dataset (26,428 echocardiographic videos), more samples would likely result in even higher performance for multi-video predictions. In contrast, the performance of the full EHR data-only model, which was consistently less than the full EHR plus videos model, was beginning to plateau. Hence, our novel multi-modal DNN approach, inclusive of echocardiography videos, provides enhanced performance for this clinical prediction task compared to what can be achieved using EHR data alone (inclusive of hand-crafted features derived by humans from the videos).

|  | NO VIDEO (~500K SAMPLES) | SINGLE VIDEO (~22K SAMPLES) | ALL VIDEOS (~27K SAMPLES) |
| --- | --- | --- | --- |
| **NO EHR VARIABLES** | 0.532 | 0.801 | 0.839 |
| **LIMITED EHR SET** | 0.786 | 0.824 | 0.843 |
| **FULL EHR SET** | 0.851 | 0.825 | **0.858** |

*Table 1: AUC scores for each data modality combination of EHR and Echo video data on the 600 left out studies used to compare to the cardiologists. "No video" models were trained on all available studies, whereas "Single Video" and "All Videos" were trained on a subset where video data were available. The No EHR variables and No Video cell denotes a random guess.*

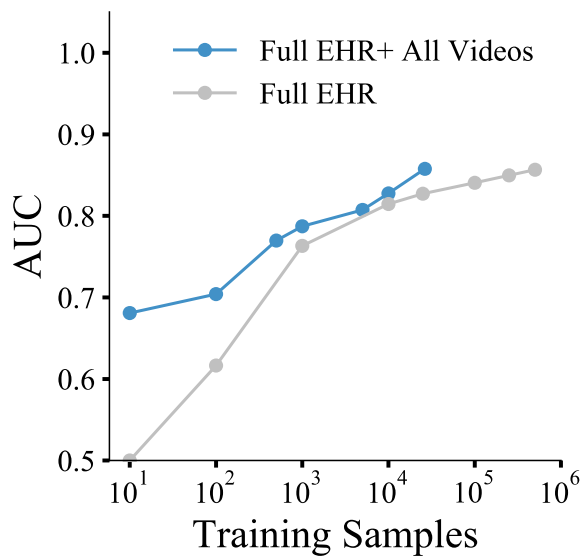

*Figure 5: Learning curves for the full (158) EHR variables model compared to the full EHR variables plus videos. The AUC is reported on the 600 patient set as a function of training set*

*size, ranging from 10 – the maximum number of datasets available for the given data inputs, which was 501,449 for the EHR variables and 26,428 for the Full EHR+videos.*

Here we demonstrated the potential for DNNs to help cardiologists predict a clinically relevant endpoint, mortality after echocardiography, using both raw video data and relevant clinical data extracted from the electronic health record. For training the DNN, we leveraged a massive dataset of 723,754 clinically-acquired videos of the heart consisting of ~45 million images. We showed that the ability of our DNN to discriminate 1-year survival—even with limited model inputs—surpassed that of trained cardiologists, suggesting that these models can add value beyond a standard clinical interpretation. To our knowledge, no prior study has demonstrated the ability to train a deep neural network to predict a future clinically-relevant event directly from image pixel-level data. Additional experiments demonstrated opportunities to achieve further significant performance gains by incorporating more EHR variables, simultaneously using all echocardiography views, and leveraging more data for model training.

We chose 1-year all-cause mortality as a highly important, easily measured clinical outcome to demonstrate feasibility for this initial work. Importantly, all-cause mortality is a well-defined endpoint without the bias that can be introduced into endpoints such as cardiovascular-specific mortality, and it can easily be extracted from an EHR that is validated against national death index databases. Moreover, mortality prediction is highly relevant for numerous applications in cardiology, as evidenced by the multitude of clinical risk scores that are currently used clinically (Framingham[27], TIMI[28], and GRACE[29] scores, etc). Future research will be needed to evaluate the performance of these models to predict additional clinically relevant outcomes in cardiology, such as hospitalizations or the need for major procedures such as a valve replacement.

Though these data had inherent heterogeneity since they were derived from a large regional healthcare system with over 10 hospitals and hundreds of clinics, additional data from other independent healthcare systems will be required to assess generalizability. Future work should be able to further improve accuracy by combining multiple videos into a single model, including Doppler based videos. Thus, methodology and architecture have been developed while feasibility and significant potential have been demonstrated for extracting predictive information from medical videos. With the ongoing rate of technological advancement and the rapid growth in electronic clinical datasets available for training, neural networks will augment future medical image interpretations with accurate predictions of clinical outcomes.

**METHODS**

**Datasets and patients.**

This retrospective study was approved by the Geisinger's Institutional Review Board and was performed with a waiver of consent.

**Image Collection and Preprocessing.**

An echocardiography study consists of several videos containing multiple views of the heart. Two clinical databases, Philips iSite and Xcelera, contained all echocardiograms collected at Geisinger. We used DCM4CHEE (version 2.0.29) and AcuoMed (version 6.0) software to retrieve a DICOM file for each echocardiography video.

The retrieved DICOM files contained an annotated video (for example, which was marked with the view name) and a raw video when the equipment was configured to store it. Without loss of generality, we used raw videos for all analyses. The raw video contained only the beam-formed ultrasound image stored in a stream of bytes format (see Extended Data Figure 1), whereas the annotated video contained artificial annotations on top of the raw video we linearly interpolated all raw videos to 30 frames per second.

Along with the video data, the DICOM file included tags that labelled the view as to which specific image orientation was acquired. These view tags had slight variations across studies for the same type of view. For example, an apical four chamber view could be tagged as "a4", "a4 2d", or "ap4". We visually inspected samples of each unique tag and grouped them into 30 common views (Extended Data Table 1). Since each video from a view group could potentially have different dimensions, we normalized all videos from a view to the most common row and column dimensions. We cropped/padded each frame with zeros to match the most common dimensions among the view group. We ultimately retrieved Philips-generated DICOM files with raw videos, view labels and excluded any videos that lasted less than 1 second.

**Electronic health record data preprocessing.**

The EHR contained 594,862 echocardiogram studies from 272,280 unique patients performed over 19 years (February 1998 to September 2018). For each study, we extracted automatic and physician reported echocardiography measurements (n = 480) along with patient demographic (n = 3), vitals (n = 5), laboratory (n = 2), and billing claims data (n = 90; International Classification of Diseases, Tenth Revision (ICD-10) ,codes from patient problem lists). For measurements taken outside of the Echocardiography study, such as fasting LDL, HDL, blood pressure, heart rate, and weight and height measurements, we retrieved the closest (before or after) within a six-month window.

All continuous variables were cleaned from physiologically out of limit values, which may have been caused by input errors. In cases where no limits could be defined for a measurement, we removed extreme outliers that met two rules: 1) Value beyond the mean plus or minus three standard deviations and 2) Value below the 25$^{th}$ percentile minus 3 interquartile ranges or above the 75$^{th}$ percentile plus 3 interquartile ranges. The removed outlier values were set as missing.

We imputed the missing data from continuous variables in two steps. First, we conducted a time interpolation to fill in missing measurements using all available studies of an individual patient, i.e., missing values in between echocardiography sessions were linearly interpolated if complete values were found in the adjacent echocardiograms. Then, to conduct Multiple Imputation by Chained Equations[30] (MICE) and complete the entire dataset, we kept 115 of 480 echocardiography measurement variables with more than 10% non-missing measurements.

We coded the reported diastolic function in an ordinal fashion with -1 for normal, 0 for dysfunction (but no grade reported), and 1, 2 and 3 for diastolic dysfunction grades I, II, and III respectively. After imputation of the continuous measurements, we imputed the missing diastolic function assessment by training a logistic regression classifier to predict the dysfunction grade (- 1, 1, 2, or 3) in a One-vs-All classifier framework using 278,160 studies where diastolic function was known.

Following imputation, we retained the physician reported left ventricular ejection fraction (LVEF) plus 57 other independent, non-redundant echocardiography measurements (i.e., excluding variables derived from other measurements; n = 58 echocardiography measurements in total).

We calculated the patient's age and survival time from the date of the echocardiogram. The patient status (dead/alive) was based on the last known living encounter or confirmed death date, which is regularly checked against national databases in our system.

We present a list and description of all 158 EHR variables used in the proposed models in the **Error! Reference source not found.**.

**Data pruning.**

The image collection and preprocessing resulted in 723,754 videos from 31,874 studies performed on 27,028 patients (an average of 22.7 videos per study). We linked the imaging and EHR data and discarded any imaging without EHR data. For a given survival experiment (3, 6, 9, and 12 months), we also removed studies without enough follow up. After that, we kept a single study per patient by randomly sampling one study per patient. This ensured that images from a single patient would not appear multiple times throughout training, validation, and testing groups.

We needed at least 600 patients (300 alive, 300 deceased), as indicated by a sample size calculation using the Pearson Chi-square test, to estimate and compare prognostic accuracy between the model and the two cardiologists. We assumed a 10% difference in accuracy between machine and cardiologist (80% vs 70%), 80% power, a significance level of 5%, and an approximate 40% discordancy. This was calculated using Power Analysis Software (PASS v15). Thus, we randomly sampled 300 studies of patients that survived and 300 that died within the set experiment threshold for each view, and set these aside from the valid samples to later compare the performance of the machine against two independent cardiologists. Only the parasternal long axis view (representing the best performing model and the cardiologists' preference for the most comprehensive single view) was ultimately used for the cardiologist comparison. The total number of valid samples for each experiment and view is shown in Extended Data Table 3, and Extended Data Figure 2.

We excluded parasternal long mitral valve, parasternal long pulmonic valve, short axis apex zoom, short axis mid papillary zoom, parasternal long lax, apical 3 zoom, and apical 2 zoom views, as they did not have enough available samples to run the experiments.

**Model selection.**

For Echocardiography video classification, we explored four different architectures: 1) A time-distributed two-dimensional Convolutional Neural Network (2D CNN) with Long Short-Term Memory (LSTM), 2) a time-distributed 2D CNN with Global Average Pooling (GAP), 3) a 3D CNN and 4) a 3D CNN with GAP. For simplicity, we abbreviate the four candidate architectures: 2D CNN + LSTM, 2D CNN + GAP, 3D CNN, and 3D CNN + GAP.

The 2D CNN + LSTM consisted of a 2D CNN branch distributed to all frames of the video. This architecture was used for a video description problem[31], where all frames from a video belonged to the same scene or action. Since all frames of the echocardiography video belong to the same scene or view, it is correct to assume that the static features would be commonly found by the same 2D kernels across the video. This assumption was put in practice for echocardiography view classification[32]. The LSTM layer aggregates the CNN features over time to output a vector that represents the entire sequence.

The 2D CNN + GAP approach exchanged the LSTM layers for the average CNN features as a time aggregation of frames. The GAP layer provides two advantages. It requires no trainable parameters, saving 1008 parameters from the LSTM layers, and enables feature interpretation. The final fully connected layer after the GAP would provide a weighted average of the CNN features, which could indicate what sections of the video weighted more in the final decision.

The 3D CNN approach aggregates time and space features as the input data flows through the network. 3D CNNs have also shown successful applications for video classification[5]. As opposed to the 2D CNN approach, 3D CNN incorporates information from adjacent frames at every layer, extracting time-space dependent features.

The 3D CNN approach would replace the Flatten operation for a GAP layer. In a similar fashion to the 2D CNN + GAP approach, the GAP layer would reduce the number of input features to the final Dense layer, thus the reduction of the number of parameters from 641 to 17; while enabling the traceback of the contributions of video features.

We defined the convolutional units of the 2D and 3D CNNs as a sequence of 7 layers in the following composition: CNN layer, Batch Normalization, ReLU, CNN layer, Batch

Normalization, ReLU, and Max Pooling (see Figure 1). All kernel dimensions were set to 3 and Max Pooling was applied in a 3 x 3 window for 2D kernels and 3 x 3 x 3 for 3D kernels.

A detailed description of the number of parameters for the 2D CNN + LSTM architecture is shown in Extended Data Table 4, 2D CNN + GAP is shown in Extended Data Table 5, 3D CNN is shown in Extended Data Table 6, and 3D CNN + GAP is shown in Extended Data Table 7. We applied all four candidate architectures to all the identified echocardiography views with a 1-year mortality label, and the 3D CNN showed consistently the best performance (Extended Data Figure 3).

Similarly, we assessed the performance gain at different image resolutions. We reduced the video resolution by factors of 2, 3, and 4. No consistent significant loss in performance was observed across all views (Extended Data Figure 4). Thus, we decided to conduct all experiments with a resolution reduction by a factor of 4 to reduce computational cost.

To incorporate EHR data into the prediction, we trained a three-layer multi-layer perceptron (MLP) with 10 hidden units at each layer. Then, we concatenated the last 10 hidden units with the CNN branch (see Figure 1).

**Training algorithm.**

We used the RMSProp[33] algorithm to train the networks with LSTM coupling, and AdaGrad[34] for the 3D CNN architectures. Each iteration of the 5-fold cross validation contained a training, validation, and test set. The training and test sets were sampled such that they had the same prevalence of alive patients, but the validation set was sampled with a balanced proportion. The validation set comprised 10% of the training set.

As we trained the DNN, we evaluated the loss (binary cross-entropy) on the validation set at each epoch. If the validation loss did not decrease for more than 10 epochs we stopped the training and reported the performance, in AUC, of the test set. We set the maximum number of epochs to 1000 and kept the default training parameters as defined by the software Keras (version 2.2). Training always ended before the maximum number of epochs was reached.

Since the prevalence of each patient class is imbalanced (~16% deceased patients), we set the weights for each class as follows:

$$w_i = \frac{Total\ Number\ of\ Samples}{2(Number\ of\ Samples\ in\ class\ i)}$$

All training was performed in an NVIDIA DGX1 platform. We independently fit each fold on each of the 8 available GPUs. The main experiment, shown in Figure 2, took a total of six days to complete.

**Effect of adding optical flow inputs.**

Optical flow velocity maps have been shown to be informative along with the original videos for classification tasks[26]. Thus, we computed the dense optical flow vectors of the echocardiography raw videos using the Gunnar Farneback's algorithm as implemented in the OpenCV (version 2.4.13.7) software library. We set the pyramid scale to 0.5, the number of levels to 3, and the window size to 5 pixels. The vectors were then converted to color videos where the color indicated direction (as in the HSV color space) and the brightness denoted amplitude. This resulted in an image video that was fed to the neural network model through an independent 3D CNN branch along with the raw video. As seen in Extended Data Figure 5, this combination of the optical flow video to the raw video did not yield consistently improved model performance compared with models using the raw video alone. Therefore, we did not use optical flow for the final study analyses.

**Use of balanced outcomes in the cardiologist survey dataset.** The 600-patient survey used to compare the accuracies of the cardiologists and the model, as described in the data pruning section, was intentionally balanced with respect to mortality outcomes (300 dead and 300 alive at one year) in order to ensure adequate power to detect differences in performance. The cardiologists were blinded to this distribution at the time of the review. We acknowledge that this balance is not reflective of typical clinical outcomes, particularly in a primary or secondary care setting, in which the base rate for 1-year survival is much higher. Hence, we cannot claim that this survey comparison between cardiologists and the model, as implemented, represents prediction in a realistic clinical setting. We do note, however, that the realistic clinical survival base rate was represented in the model training/testing sets, just as in the conditioning experiences of the cardiologists (consistent with their preference—high specificity for death—in over-estimating 1-year survival). Thus, the model was not advantaged in this regard by learning to expect this different outcome. Instead, rather than prediction informed by clinical base rates, our comparison sought to evaluate the true discriminative abilities and accuracies of the cardiologists compared to the machine.

**Software for cardiologist survey.** We deployed a web application with the interface shown in Extended Data Figure 6. The application required the cardiologist to input their institutional credentials for access. We showed the 10 EHR variables and the two versions of the video, raw and annotated. The application then recorded the cardiologist prediction as they clicked on either the "Alive" or "Dead" buttons.

**Statistical analysis of comparison between Machine and Cardiologists.** The cardiologists' responses were binary, and the Machine's response was continuous. We set 0.5 as the threshold for the Machine's response prior to performing the final comparison experiment. Since all responses were recorded for the same samples, we conducted a Cochran's Q test to assess whether the three responses where significantly different in the proportion of correctly classified samples. This test showed that there was enough evidence that at least one of the responses was significantly different with a p-value of 1.8e-15. A post hoc analysis of pairwise comparisons between the three responses resulted in Bonferroni-adjusted p-values of 0.003, $4.2e^{-11}$, and $6.9e^{-7}$ for the pairs Cardiologist 1 vs Cardiologist 2, Cardiologist 1 vs Machine, and Cardiologist 2 vs Machine, respectively.

**Use of human subjects.** This Human Subjects Research falls under Exemption 4 of the Health and Human Services human subject regulations since the research was conducted on existing patient data from the electronic health record at our institution.

**Data availability statement.** The medical training / validation data which were used for the current study are the property of Geisinger and are not publicly available due to the presence of patient identifiers. Some data may be available from the authors upon reasonable request and with permission from Geisinger.

Extended Data Table 1: View labels found in DICOM tags for the corresponding view type. The view tag in bold indicates the abbreviation used for the view type.

| VIEW TYPE | VIEW TAG |
|---|---|
| APICAL 2 | **a2**, ap2 2d, a2 2d, a2 lavol, la 2ch |
| APICAL 3 | **a long**, ap3 2d, a3 2d |
| APICAL 4 | **ap4**, ap4 2d, a4 2d, a4 zoom, a4 lavol, la ap4 ch |
| APICAL 4 FOCUSED TO RV | **rv focus**, rvfocus |
| APICAL 5 | **a5**, ap5 2d, a5 2d |
| PARASTERNAL LONG AXIS | **pl deep**, psl deep |
| PARASTERNAL LONG ASCENDING AORTA | **pl ascao**, asc ao, pl asc ao |
| PARASTERNAL LONG MITRAL VALVE | **pla mv** |
| PARASTERNAL LONG PULMONIC VALVE | **pl pv**, pv lax |
| PARASTERNAL LONG RV INFLOW | **pl rvif**, rv inf, rvif 2d |
| PARASTERNAL LONG ZOOM AORTIC VALVE | **pl av ao**, av zoom |
| PARASTERNAL SHORT AORTIC VALVE | **ps av**, psavzoom, psax av |
| PARASTERNAL SHORT PULMONIC VALVE AND PULMONARY ARTERY | **ps pv pa**, ps pv, psax pv |
| PARASTERNAL SHORT TRICUSPID VALVE | **ps tv**, ps tv 2d, psax tv |
| SHORT AXIS APEX | **sax apex** |
| SHORT AXIS BASE | **lv base** |
| SHORT AXIS MID PAPILLARY | sax mid, sax |
| SUBCOSTAL 4CHAMBER | **sbc 4 ch**, sbc 4, sbc 4ch |
| SUBCOSTAL HEPATIC VEIN | **ivc hv**, sbc hv |
| SUBCOSTAL INTER-ATRIAL SEPTUM | **ias**, sbc ias, ias 2d |
| SUBCOSTAL IVC WITH RESPIRATION | **ivc resp**, sbc ivc, ivc insp, ivc snif, ivcsniff, sniff |
| SUBCOSTAL RV | **sbc rv** |
| SUPRASTERNAL NOTCH | **ssn**, ssn sax |
| PARASTERNAL LONG LAX | lax |
| SHORT AXIS MID PAPILLARY | **lv mid** |
| SHORT AXIS APEX | **lv apex** |
| APICAL 3 ZOOM | ap3 |
| APICAL 2 ZOOM | ap2 |
| SHORT AXIS BASE | sax base |

Extended Data Table 2: Description of all variables extracted from the electronic health records. *MOD = modified ellipsoid, **el = (single plane) ellipsoid, LV = left ventricular, IV = inter-ventricular. [1—10] Selected EHR variables previously reported as the top 10 predictors of 1-year mortality.

|  | EHR VARIABLE | UNITS | VARIABLE CLASS | DESCRIPTION |
|---|---|---|---|---|
| 1 | Age[1] | years | demographics | At the time of Echocardiography study |
| 2 | Sex | 0: Female, 1: Male | demographics | |
| 3 | Smoking status | 0: No, 1: Yes | demographics | Ever smoked |
| 4 | Height | cm | vitals | |
| 5 | Weight | kg | vitals | |
| 6 | Heart rate[3] | bpm | vitals | |
| 7 | Diastolic blood pressure[6] | mm Hg | vitals | |
| 8 | Systolic blood pressure[8] | mm Hg | vitals | |
| 9 | LDL[4] | mg/DL | laboratory | Low-density lipoprotein |
| 10 | HDL | mg/DL | laboratory | High-density lipoprotein |
| 11 | LVEF[5] | % | Echo measure | Physician-reported left ventricular ejection fraction |
| 12 | AI dec slope | cm/s2 | Echo measure | Aortic insufficiency deceleration slope |
| 13 | AI max vel | cm/s | Echo measure | Aortic insufficiency maximum velocity |
| 14 | Ao V2 VTI | cm | Echo measure | Velocity-time integral of distal to aortic valve flow |
| 15 | Ao V2 max | cm/s | Echo measure | Maximum velocity of distal to aortic valve flow |
| 16 | Ao V2 mean | cm/s | Echo measure | Mean velocity of distal to aortic valve flow |
| 17 | Ao root diam | cm | Echo measure | Aortic root diameter |
| 18 | Asc Aorta | cm | Echo measure | Ascending aortic diameter |
| 19 | EDV MOD*-sp2 | ml | Echo measure | LV end-diastolic volume: apical 2-chamber |
| 20 | EDV MOD*-sp4 | ml | Echo measure | LV end-diastolic volume: apical 4-chamber |
| 21 | EDV sp2-el** | ml | Echo measure | LV end-diastolic volume: apical 2-chamber |
| 22 | EDV sp4-el** | ml | Echo measure | LV end-diastolic volume: apical 4-chamber |
| 23 | ESV MOD*-sp2 | ml | Echo measure | LV end-systolic volume: apical 2-chamber |
| 24 | ESV MOD*-sp4 | ml | Echo measure | LV end-systolic volume: apical 4-chamber |
| 25 | ESV sp2-el** | ml | Echo measure | LV end-systolic volume: apical 2-chamber |
| 26 | ESV sp4-el** | ml | Echo measure | LV end-systolic volume: apical 4-chamber |
| 27 | IVSd | cm | Echo measure | IV septum dimension at end-diastole |
| 28 | LA dimension | cm | Echo measure | Left atrium dimension |
| 29 | LAV MOD*-sp2 | ml | Echo measure | Left atrium volume: apical 2-chamber |
| 30 | LAV MOD*-sp4 | ml | Echo measure | Left atrium volume: apical 4-chamber |
| 31 | LV V1 VTI | cm | Echo measure | Velocity-time integral: proximal to the obstruction |
| 32 | LV V1 max | cm/s | Echo measure | Maximum LV velocity: proximal to the obstruction |
| 33 | LV V1 mean | cm/s | Echo measure | Mean LV velocity proximal to the obstruction |
| 34 | LVAd ap2 | cm2 | Echo measure | LV area at end-diastole: apical 2-chamber |
| 35 | LVAd ap4 | cm2 | Echo measure | LV area at end-diastole: apical 4-chamber |
| 36 | LVAs ap2 | cm2 | Echo measure | LV area at end-systole: apical 2-chamber |
| 37 | LVAs ap4 | cm2 | Echo measure | LV area at end-systole: apical 4-chamber |
| 38 | LVIDd | cm | Echo measure | LV internal dimension at end-diastole |
| 39 | LVIDs | cm | Echo measure | LV internal dimension at end-systole |
| 40 | LVLd ap2 | cm | Echo measure | LV long-axis length at end-diastole: apical 2-chamber |
| 41 | LVLd ap4 | cm | Echo measure | LV long-axis length at end-diastole: apical 4-chamber |
| 42 | LVLs ap2 | cm | Echo measure | LV long-axis length at end systole: apical 2-chamber |
| 43 | LVLs ap4 | cm | Echo measure | LV long-axis length at end systole: apical 4-chamber |
| 44 | LVOT area M | cm2 | Echo measure | LV outflow tract area |
| 45 | LVOT diam | cm | Echo measure | LV outflow tract diameter |
| 46 | LVPWd | cm | Echo measure | LV posterior wall thickness at end-diastole |
| 47 | MR max vel | cm/s | Echo measure | Mitral regurgitation maximum velocity |
| 48 | MV A point | cm/s | Echo measure | A-point maximum velocity of mitral flow |
| 49 | MV E point | cm/s | Echo measure | E-point maximum velocity of mitral flow |
| 50 | MV P1/2t max-vel | cm/s | Echo measure | Maximum velocity of mitral valve flow |
| 51 | MV dec slope | cm/s2 | Echo measure | Mitral valve deceleration slope |
| 52 | MV dec time | s | Echo measure | Mitral valve deceleration time |
| 53 | PA V2 max | cm/s | Echo measure | Maximum velocity of distal to pulmonic valve flow |
| 54 | PA acc slope[9] | cm/s2 | Echo measure | Pulmonary artery acceleration slope |
| 55 | PA acc time[7] | s | Echo measure | Pulmonary artery acceleration time |
| 56 | Pulm. R-R | s | Echo measure | Pulmonary R-R time interval |
| 57 | RAP systole | mm-Hg | Echo measure | Right atrial end-systolic mean pressure |
| 58 | RVDd | cm | Echo measure | Right ventricle dimension at end-diastole |
| 59 | TR max vel[2] | cm/s | Echo measure | Tricuspid regurgitation maximum velocity |
| 60 | AVR | 0/1 Hot encoded for severity levels 0,1,2,3 | Echo measure | Aortic valve regurgitation |
| 61 | MVR | 0/1 Hot encoded for severity levels 0,1,2,3 | Echo measure | Mitral valve regurgitation |
| 62 | TVR | 0/1 Hot encoded for severity levels 0,1,2,3 | Echo measure | Tricuspid valve regurgitation |
| 63 | PVR | 0/1 Hot encoded for severity levels 0,1,2,3 | Echo measure | Pulmonary valve regurgitation |
| 64 | AVS | 0/1 Hot encoded for severity levels 0,1,2,3 | Echo measure | Aortic valve stenosis |
| 65 | MVS | 0/1 Hot encoded for severity levels 0,1,2,3 | Echo measure | Mitral valve stenosis |
| 66 | TVS | 0/1 Hot encoded for severity levels 0,1,2,3 | Echo measure | Tricuspid valve stenosis |
| 67 | PVS | 0/1 Hot encoded for severity levels 0,1,2,3 | Echo measure | Pulmonary valve stenosis |
| 68 | Diastolic function[10] | -1: Normal, 0: abnormal (no grade reported), [1,2,3]: grade I/II/II | Echo measure | Physician-reported diastolic function |
| 69 – 71 | I00, I01, I02 | | Diagnosis code | Acute rheumatic fever |
| 72 – 76 | I05, I06, I07, I08, I09 | | Diagnosis code | Chronic rheumatic heart disease |
| 77 – 82 | I10, I11, I12, I13, I15, I16 | | Diagnosis code | Hypertensive diseases |
| 83 – 88 | I20, I21, I22, I23, I24, I25 | | Diagnosis code | Ischemic heart diseases |
| 89 – 91 | I26, I27, I28 | | Diagnosis code | Pulmonary heart disease and diseases of pulmonary circulation |
| 92 | I30 | | Diagnosis code | Acute pericarditis |
| 93 – 106 | I31, I32, I33, I34, I35, I36, I37, I38, I39, I43, I44, I45, I49, I51 | | Diagnosis code | Other forms of heart disease |
| 107 | I40 | | Diagnosis code | Acute myocarditis |
| 108 | I42 | | Diagnosis code | Cardiomyopathy |
| 109 | I46 | | Diagnosis code | Cardiac arrest |
| 110 | I47 | | Diagnosis code | Paroxysmal tachycardia |
| 111 | I48 | | Diagnosis code | Atrial fibrillation |
| 112 | I50 | | Diagnosis code | Heart failure |
| 113 – 121 | I60, I61, I62, I63, I65, I66, I67, I68, I69 | | Diagnosis code | Cerebrovascular diseases |
| 122 – 131 | I70, I71, I72, I73, I74, I75, I76, I77, I78, I79 | | Diagnosis code | Diseases of arteries, arterioles and capillaries |
| 131 – 140 | I80, I81, I82, I83, I85, I86, I87, I88, I89 | | Diagnosis code | Diseases of veins, lymphatic vessels, and lymph nodes |
| 141 | I95 | | Diagnosis code | Hypotension |
| 142 – 144 | I96, I97, I99 | | Diagnosis code | Other and unspecified disorders of the circulatory system |
| 145 – 149 | E08, E09, E10, E11, E13 | | Diagnosis code | Diabetes mellitus |
| 150 – 156 | Q20, Q21, Q22, Q23, Q24, Q25, Q26 | | Diagnosis code | Congenital heart defect |
| 157 | E78 | | Diagnosis code | Dyslipidemia |
| 158 | N18 | | Diagnosis code | Chronic kidney disease |

Extended Data Table 3: Number of valid samples after setting 600 studies aside for the final test comparison to the 2 cardiologists.

| VIEW GROUP | 3 MONTHS | 6 MONTHS | 9 MONTHS | 12 MONTHS |
|---|---|---|---|---|
| APICAL 2 | 19,334 | 19,328 | 19,323 | 19,316 |
| APICAL 3 | 19,392 | 19,388 | 19,384 | 19,376 |
| APICAL 4 | 18,755 | 18,749 | 18,745 | 18,737 |
| APICAL 4 FOCUSED TO RV | 21,192 | 21,186 | 21,181 | 21,173 |
| APICAL 5 | 18,438 | 18,431 | 18,426 | 18,419 |
| PARASTERNAL LONG AXIS | 22,426 | 22,420 | 22,415 | 22,407 |
| PARASTERNAL LONG ASCENDING AORTA | 21,700 | 21,694 | 21,688 | 21,681 |
| PARASTERNAL LONG RV INFLOW | 21,544 | 21,538 | 21,534 | 21,528 |
| PARASTERNAL LONG ZOOM AORTIC VALVE | 21,657 | 21,650 | 21,645 | 21,637 |
| PARASTERNAL SHORT AORTIC VALVE | 21,875 | 21,870 | 21,865 | 21,857 |
| PARASTERNAL SHORT PULMONIC VALVE AND PULMONARY ARTERY | 21,614 | 21,609 | 21,605 | 21,596 |
| PARASTERNAL SHORT TRICUSPID VALVE | 13,385 | 13,379 | 13,375 | 13,370 |
| SHORT AXIS BASE | 21,541 | 21,535 | 21,530 | 21,523 |
| SUBCOSTAL 4 CHAMBER | 20,768 | 20,763 | 20,758 | 20,751 |
| SUBCOSTAL HEPATIC VEIN | 11,033 | 11,029 | 11,024 | 11,020 |
| SUBCOSTAL INTER-ATRIAL SEPTUM | 19,402 | 19,399 | 19,394 | 19,387 |
| SUBCOSTAL IVC WITH RESPIRATION | 20,510 | 20,505 | 20,499 | 20,492 |
| SUBCOSTAL RV | 20,263 | 20,259 | 20,254 | 20,247 |
| SUPRASTERNAL NOTCH | 18,382 | 18,378 | 18,372 | 18,365 |
| SHORT AXIS MID PAPILLARY | 21,801 | 21,796 | 21,791 | 21,783 |
| SHORT AXIS APEX | 21,870 | 21,864 | 21,859 | 21,851 |

Extended Data Table 4: Time-distributed 2D Convolutional Neural Network with Long Short-Term Memory aggregation (2D CNN + LSTM). Number of parameters in Conv layers in the format CNN + Batch Normalization + CNN + Batch Normalization = Total number of parameters combined

| LAYER NAME | INPUT DIMENSIONS | NUMBER OF PARAMETERS |
|---|---|---|
| TIME-DISTRIBUTED 2D CONV 1 | 60x109x150x1 | 40+16+148+16= 220 |
| TIME-DISTRIBUTED 2D CONV 2 | 60x36x50x4 | 296+32+584+32=944 |
| TIME-DISTRIBUTED 2D CONV 3 | 60x12x16x8 | 1,168+64+2,320+64=3,616 |
| TIME-DISTRIBUTED 2D CONV 4 | 60x4x5x16 | 2,320+64+2,320+64=4,768 |
| TIME-DISTRIBUTED FLATTEN | 60x1x1x16 | 0 |
| LSTM 1 | 60x16 | 800 |
| LSTM 2 | 60x8 | 208 |
| DENSE | 4 | 5 |
|  | Total | 10,561 |

Extended Data Table 5: Time-distributed 2D Convolutional Neural Network with Global Average Pooling aggregation (2D CNN + GAP).

| LAYER NAME | INPUT DIMENSIONS | NUMBER OF PARAMETERS |
|---|---|---|
| **TIME-DISTRIBUTED 2D CONV 1** | 60x109x150x1 | 40+16+148+16= 220 |
| **TIME-DISTRIBUTED 2D CONV 2** | 60x36x50x4 | 296+32+584+32=944 |
| **TIME-DISTRIBUTED 2D CONV 3** | 60x12x16x8 | 1,168+64+2,320+64=3,616 |
| **TIME-DISTRIBUTED 2D CONV 4** | 60x4x5x16 | 2,320+64+2,320+64=4,768 |
| **GLOBAL AVERAGE POOLING** | 60x4x5x16 | 0 |
| **DENSE** | 16 | 17 |
| | Total | 9,565 |

Extended Data Table 6: 3D Convolutional Neural Network with Global Average Pooling aggregation (3D CNN + GAP).

| LAYER NAME | FEATURE DIMENSIONS | NUMBER OF PARAMETERS |
| --- | --- | --- |
| 3D CONV 1 | 60x109x150x1 | 112+16+436+16=580 |
| 3D CONV 2 | 20x36x50x4 | 872+32+1,736+32=2672 |
| 3D CONV 3 | 6x12x16x8 | 3,472 +64+6,928+64=10,528 |
| GLOBAL AVERAGE POOLING | 6x12x16x16 | 0 |
| DENSE | 16 | 17 |
| | Total | 13,797 |

Extended Data Table 7: 3D Convolutional Neural Network (3D CNN).

| LAYER NAME | FEATURE DIMENSIONS | NUMBER OF PARAMETERS |
|---|---|---|
| 3D CONV 1 | 60x109x150x1 | 112+16+436+16=580 |
| 3D CONV 2 | 20x36x50x4 | 872+32+1,736+32=2672 |
| 3D CONV 3 | 6x12x16x8 | 3,472 +64+6,928+64=10,528 |
| FLATTEN | 2x4x5x16 | 0 |
| DENSE | 640 | 641 |
|  | Total | 14,421 |

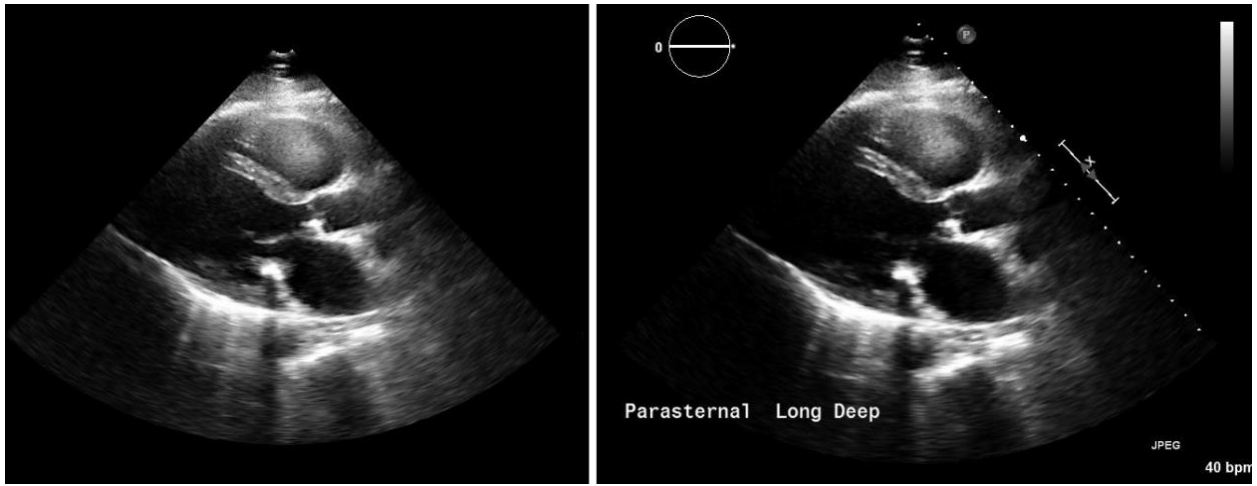

Extended Data Figure 1: Examples of raw (left) and annotated (right) videos.

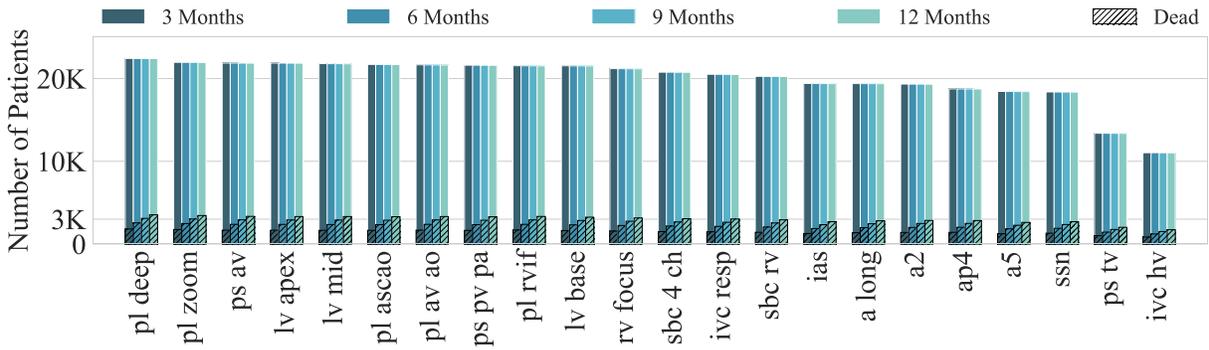

Extended Data Figure 1: Plot of the number of patients for experiments that required 3, 6, 9, and 12 months follow-up (as indicated in the Extended Data Table 2) with the proportion of dead patients (shaded bar).

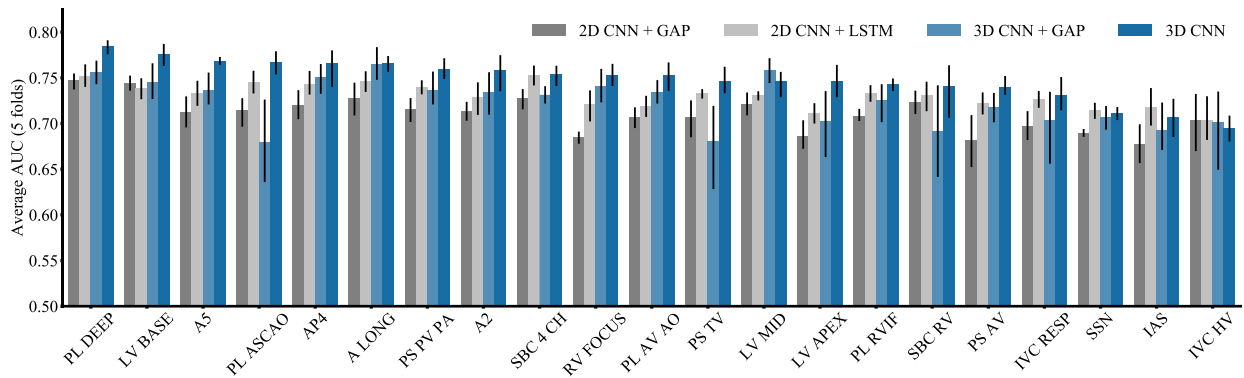

Extended Data Figure 2: AUCs of one-year mortality predictions across all views with four different neural network architectures: 2D CNN + Global Average Pooling (GAP; dark gray), 2D CNN + Long Short-Term Memory (LSTM; light gray), a 3D CNN + GAP (light blue), and 3D CNN (dark blue).

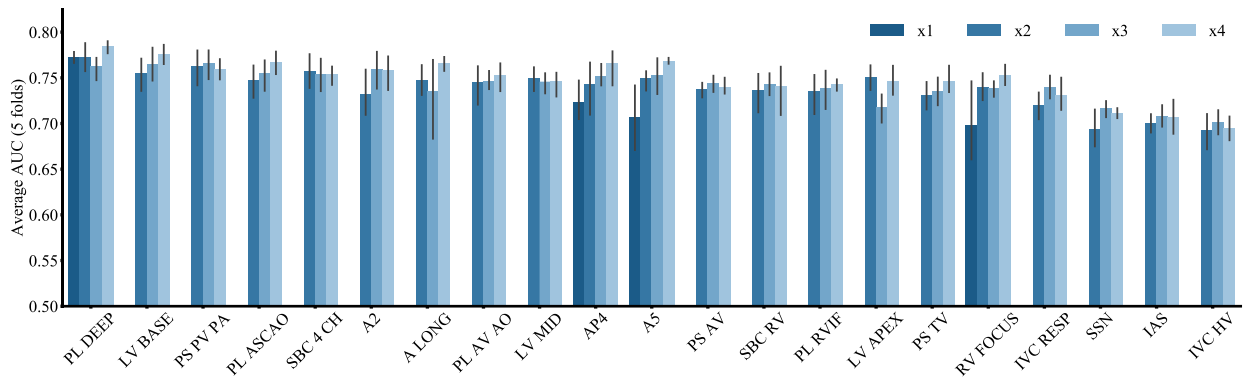

Extended Data Figure 4: AUCs of one-year mortality predictions across all views with different levels of reduced resolution ranging from native (x1) to 4-fold (x4). Note that full native resolution training was only done for select views due to the computational time required to complete the experiment at this resolution.

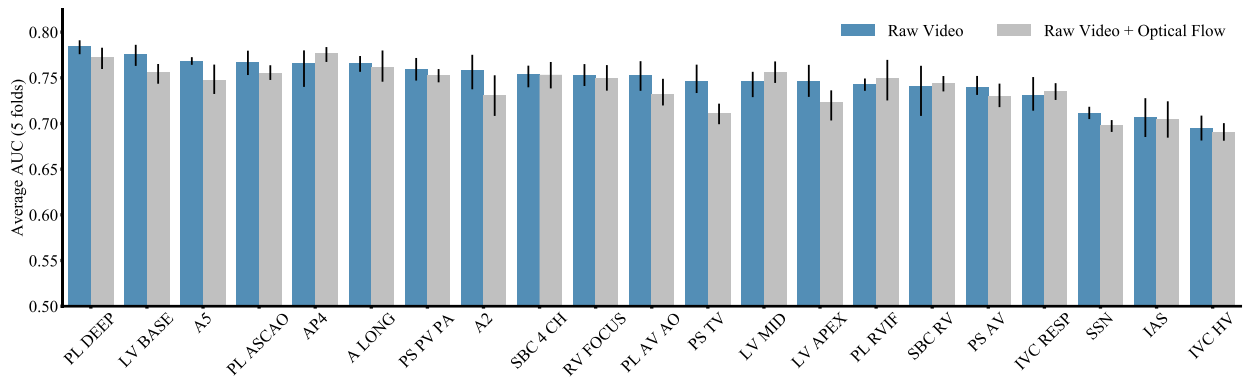

Extended Data Figure 5: One-year mortality prediction performance ranking for all echocardiography views using only the raw video (blue) versus the raw video with optical flow features (gray).

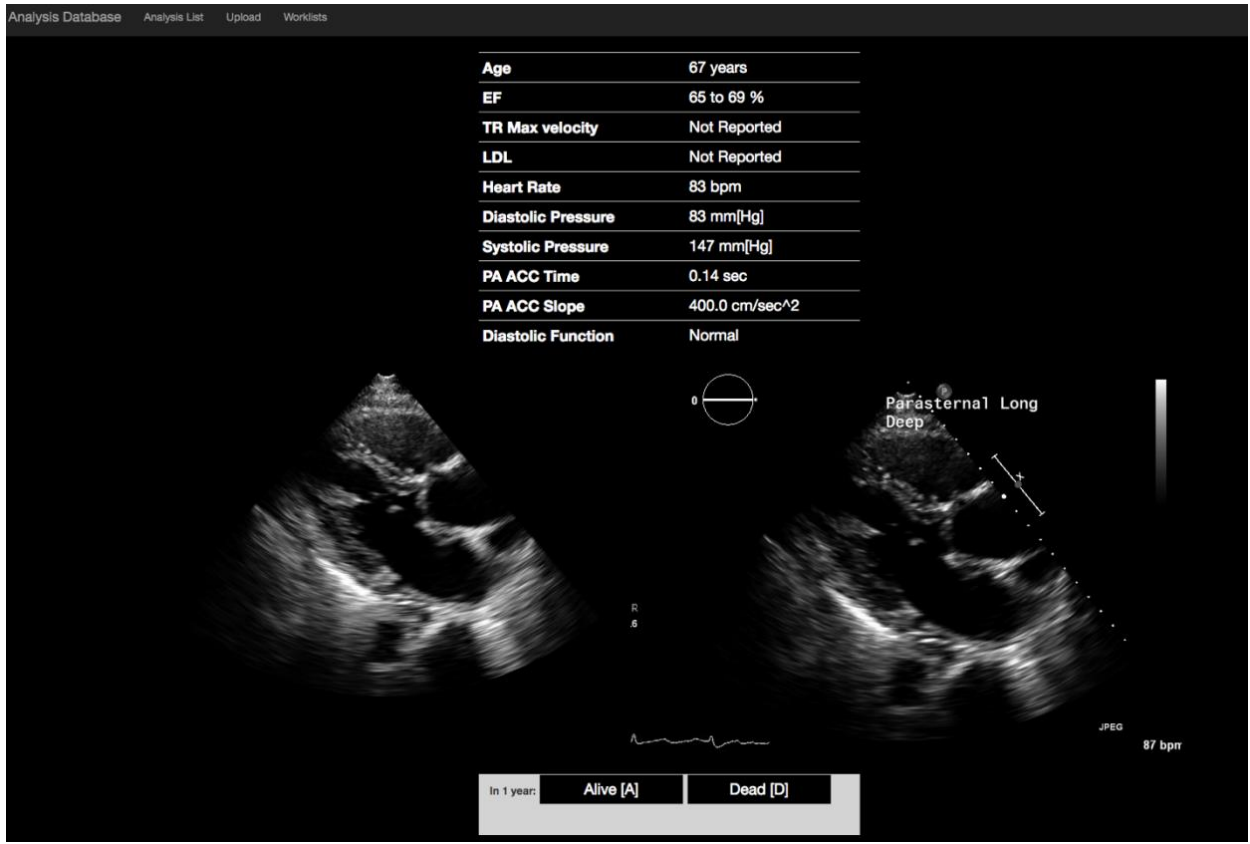

Extended Data Figure 6: Interface of the web application developed for cardiologists to predict survival one year after echocardiography.